\documentclass{article} % For LaTeX2e
\usepackage{iclr2026_conference,times}
\usepackage{xcolor}
\usepackage{amsmath,amsfonts,bm}

\def\eqref#1{equation~\ref{#1}}
\def\1{\bm{1}}

\DeclareMathAlphabet{\mathsfit}{\encodingdefault}{\sfdefault}{m}{sl}
\SetMathAlphabet{\mathsfit}{bold}{\encodingdefault}{\sfdefault}{bx}{n}

\usepackage{hyperref}
\usepackage{url}
\usepackage{comment}
\usepackage{booktabs}
\usepackage{graphicx}
\usepackage{wrapfig}

\title{Probabilistic Multivariate Time Series \\Forecasting
with Diffusion Copulas}

\author{%
  David Huk\thanks{Equal contribution.} \\
  Department of Statistics \\
  The University of Warwick \\
  \texttt{david.huk@warwick.ac.uk} \\
  \And
  Dongshan Wang\textsuperscript{*} \\ 
  Department of Statistics \\
  The University of Warwick \\
  \texttt{Dongshan.wang@warwick.ac.uk} \\ 
  \And
  Miha Bresar\thanks{Corresponding author.} \\
  School of Data Science \\
  The Chinese University of Hong Kong, Shenzhen \\
  \texttt{mihabresar@cuhk.edu.cn}
}

\iclrfinalcopy % Uncomment for camera-ready version, but NOT for submission.
\begin{document}

\maketitle

\begin{abstract}
Accurately assessing financial risk requires capturing both individual asset volatility and the complex, asymmetric dependence structures that emerge during extreme market events. While modern diffusion-based models have advanced multivariate forecasting, they often suffer from a "normality bias" when trained end-to-end—sacrificing marginal calibration for joint coherence and consistently underestimating tail risk. To address this, we propose a Diffusion-Copula framework that explicitly decouples the learning of marginal distributions from their dependence structure. We employ deep Mixture Density Networks to capture heavy-tailed asset dynamics, followed by a Classification-Diffusion Copula to model the joint dependence. Applied to cryptocurrency markets, our approach demonstrates superior performance over state-of-the-art baselines in forecasting systemic extremes of both marginal and joint events. Crucially, we demonstrate that while baseline models classify simultaneous market crashes as statistically impossible "Black Swans" (high surprise), our framework identifies them as "Expected Crashes" (low surprise), successfully preserving the correlation structure necessary for robust risk management during contagion events.
\end{abstract}

\section{Motivation}

Accurately assessing financial risk requires capturing both individual asset volatility and the joint dependence of extreme events~\citep{embrechts2013model, patton2006modelling}. Historically, statistical copula models \citep{joe2014dependence} addressed this dilemma by modelling assets and their dependence structures separately. The industry relied on fixed parametric families whose brittleness was exposed during the 2008 financial crisis. The models' critical flaw being the assumption of asymptotic tail independence~\citep{salmon2012formula, donnelly2010devil}, which contradicts the empirical reality of contagion, where correlations converge to unity during market crashes.

Recent research has pivoted toward Diffusion models as a powerful, data-driven alternative for temporal data~\citep{MR4982120, li2024transformer, tashiro2021csdi}. However, applying multivariate diffusion models directly often compromises marginal calibration, making it difficult to sample consistent extremes across assets of varying magnitudes.

To address this, we propose a Diffusion-Copula framework that reconsiders the copula approach using modern generative AI. We adopt a two-step strategy: first, marginal time series are modeled independently with light-weight LSTMs to ensure precise calibration and absorb serial dependence. Second, to reconcile these marginals and induce dependence, we employ the Classification-Diffusion Copula~\citep{huk2025diffusion}. This hybrid approach avoids the rigidity of Gaussian methods while preserving marginal fidelity, allowing us to effectively capture the joint tail events. 

We apply our model to cryptocurrency markets, demonstrating that it outperforms standard diffusion baselines~\citep{MR4982120, rasul2021autoregressive, tashiro2021csdi} in predicting joint tail events. Accurate tail modelling is of crucial importance in this context. Cryptocurrencies are not only highly volatile, but exhibit strong tail dependence during crashes, meaning they tend to collapse simultaneously~\citep{borri2019conditional, bouri2021quantile}. By explicitly capturing this synchronisation, our Diffusion-Copula approach offers a significant advantage over models that fail to account for these extreme dependencies.

\section{Background}

\subsection{Probabilistic time series forecasting}

We employ a probabilistic forecasting framework to model the stochastic dynamics of returns for each of $d$ assets, denoted by $y_t=\{y_t^1,\ldots,y_t^d\} \in\mathbb{R}^d$ for a time $t$. Our objective is to learn the \textbf{joint predictive distribution} of the one-step-ahead values $y_{t}$, conditional on a history of past observations $\mathcal{X}_t = (y_{t-k}, \dots, y_{t-1})$. Rather than producing a deterministic point estimate, we estimate the conditional probability density $p(y_t \mid \mathbf{x}_t)$ directly. This task is more challenging than pure point estimation because both the uncertainties of each individual asset and the uncertainty in the interplay between assets must be taken into account.

\subsection{Copula decomposition for multivariate forecasts}

A candidate approach for tackling this problem is found in the class of models known as \textit{copulas}. These models link together independent dimension-wise distributions, providing the dependence structure missing from a factorised model~\citep{joe2014dependence}. As established by Sklar's theorem~\citep{MR125600}, a joint probability density function can be decomposed into the product of its marginal densities and a copula density that captures the inter-variable dependence.

In the context of conditional predictive modelling, Sklar's theorem leads to the following decomposition of the joint predictive density as a product of marginals and a copula density:
\begin{equation}
    \label{eq:cond_sklar}
    {p}(\mathbf{y}_t|\mathcal{X}_t)= \prod_{i=1}^{d} \{p^i(y_t^i|\mathcal{X}_t) \}\cdot c(u_t^1,\ldots,u_t^d|\mathcal{X}_t),
\end{equation}
where $u_t^i = P^i(y_t^i|\mathcal{X}_t)$ represents the observations transformed to the copula scale via their marginal cumulative distribution functions (CDFs). Here, $c:[0,1]^d\mapsto(0,\infty)$ is the copula density supported on the unit hypercube.

While the general copula density $c$ may depend on the predictors $\mathcal{X}_t$, a common modelling choice which we adopt is to assume the dependence on predictors is fully encapsulated by the marginal distributions $P^i$. This disentanglement allows for a modular approach: one can first model the univariate variables independently to handle non-stationarity, and subsequently model their dependence using a copula on the resulting CDF values. This assumption reduces the complexity of the dependence modelling task strictly to the output dimension $d$, recovering the scalability of factorised models while maintaining valid joint dependence.

%However, the efficacy of this modular approach hinges entirely on the quality of the marginal estimators. For the copula $c$ to be valid, the inputs $u_t^i$ must be strictly uniform on $[0,1]$. This requirement places a heavy burden on the marginal models $P^i$.
%\begin{itemize}
    %\item \textbf{Calibration:} 
    %If the marginals are mis-specified, the transformed variables $u$ will not be uniform, thereby violating the fundamental assumption of the copula and distorting the learned dependence.
    %\item \textbf{Modelling Choices:} 
    For marginal models, one may employ parametric families parameterised by neural networks to absorb non-stationarity, leaving i.i.d.\ noise for the copula. This lets practitioners benefit from powerful architectures with inherent temporal biases, such as Long Short-Term Memory networks~\citep{hochreiter1997long} and attention~\citep{vaswani2017attention} layers, while also providing guarantees such as heavy tails.  Alternatively, non-parametric approaches (like splines or empirical CDFs) offer flexibility but may struggle with extrapolation in tail regions, which are critical for accurately representing extreme events.
%\end{itemize}

\section{Method}

\subsection{Preprocessing and Problem Setup}

We define the discrete time series of returns $y_t$ derived from the raw price sequence $P_t$ as the percentage change:
\begin{equation}
    y_t = \frac{P_t - P_{t-1}}{P_{t-1}}.
\end{equation}
We do this in order to preserve time-homogeneous structure of the time-series. Working with raw prices directly would force us to estimate a large ``price-drift'' arising from the fast growth of the cryptocurrencies we are studying in the present paper.

As part of our copula-based approach, the first step is correctly modelling the predictive marginal distribution $p^i(y^{i}_t \mid \mathcal{X}_t)$ for every dimension $i$. As these distributions are separate, we model each dimension independently with a specialized marginal model.

\subsection{Predictive Marginal Model}
\label{sec:pred_marg}
We approximate the conditional probability density function of the returns using a \textbf{Mixture Density Network (MDN)}. The conditional density is defined as a weighted sum of heterogeneous component distributions—specifically Normal, Laplace, and Student-t distributions—parameterized by a neural network $\Psi$:

\begin{equation}
    p^i(y_t^i \mid \mathcal{X}_t; \Psi) = \sum_{k=1}^{K} \pi_k(\mathcal{X}_t) \cdot f_k(y_t^i; \theta_k(\mathcal{X}_t))
\end{equation}

where $\pi_k(\mathcal{X}_t)$ are the mixing coefficients satisfying $\sum \pi_k = 1$, and $f_k$ represents the component densities. The parameters $\theta_k$ (location $\mu$, scale $\sigma$, and degrees of freedom $\nu$) are dynamic functions of the history $\mathcal{X}_t$. This construction enables heavy tail coverage while also offering enough flexibility to capture multimodality or skewness of marginal predictives.

\subsection{Markovian Assumption}

To maintain computational tractability while capturing short-term temporal dependencies, we employ a Markov assumption of order $k=14$. We assume that the distribution of the current return $y_t$ is conditional only on the window of the most recent 14 observations:
\begin{equation}
    \mathcal{X}_t = \{ y_{t-14}, y_{t-13}, \dots, y_{t-1} \}
\end{equation}
Consequently, the neural network input dimension is fixed at 14 time steps, allowing the model to learn localised volatility clusters and trend reversals present in the immediate history.

\paragraph{Marginal training}
The optimization loop conditions the neural network on the history $\mathcal{X}_t$, comprising $k$ lagged returns concatenated with auxiliary features such as rolling volatility and path geometry metrics (e.g., maximum drawdown, trend strength), to output the parameters $\theta$ (mixing coefficients $\pi$, locations $\mu$, scales $\sigma$, and degrees of freedom $\nu$) of our {Mixture Density Network}. This model explicitly constructs a parametric conditional density $p(y_t \mid \mathcal{X}_t)$. The parameters are updated via gradient descent to minimize the \textbf{Negative Log-Likelihood (NLL)}:

\begin{equation}
    \mathcal{L}(\theta) = -\frac{1}{N} \sum_{t=1}^{N} \log \left( \sum_{k=1}^{K} \pi_k(\mathcal{X}_t) f_k(y_t; \theta_k) \right) + \lambda \mathcal{H}(\pi)
\end{equation}

where $f_k$ are the component densities (Normal, Laplace, Student-t) and $\mathcal{H}(\pi)$ represents the entropy regularisation on mixture weights used to encourage diverse utilisation of mixture components (full architectural and training details are included in Appendix \ref{apdx:exp}).

\subsection{Copula model}
To unite marginal predictives into a joint multivariate predictive distribution, we capture the dependence of the $d$-dimensional financial time series via a copula model. We denote the marginal cumulative distribution functions by $F_i(x^i)$ (modelled via MDN as described above). We obtain copula-scale data $u \in [0, 1]^d$ via the probability integral transform $u^i = F_i(x^i)$. By Sklar's theorem \citep{MR125600}, the joint density of our model is given by: 
\begin{equation}
    {p}(\mathbf{y}_t|\mathcal{X}_t)= \prod_{i=1}^{d} \{p^i(y_t^i|\mathcal{X}_t) \}\cdot c(u_t^1,\ldots,u_t^d|\mathcal{X}_t),
\end{equation}
where $c(\cdot)$ captures the multivariate dependence structure.

 We employ the \textbf{Classification-Diffusion Copula (CDC)}~\citep{huk2025diffusion} as our model for $c(\cdot)$, utilising the strength and flexibility of diffusion models for dependence. The CDC models dependence by mapping the copula data $u$ to the Gaussian scale $z^i = \Phi^{-1}(u^i)$ and defining a forward Ornstein-Uhlenbeck (OU) process that progressively destroys the dependence structure while preserving the standard Gaussian marginals. The core insight of the method is that the copula density can be recovered by training a classifier to distinguish between "dependent" (early diffusion time $t=0$) and "independent" (late diffusion time $t=T_k$) states of this process. Specifically, the copula density is given by the ratio of class probabilities predicted by a neural network $f_\theta$:
\begin{equation}
    c(u) = \frac{\mathbb{P}_\theta(t=0 \mid z = \Phi^{-1}(u))}{\mathbb{P}_\theta(t=T_K \mid z = \Phi^{-1}(u))}
\end{equation}
where $t=0$ represents the data distribution and $T_K \to\infty$ represents the independent noise distribution (this follows from Bayes' rule on the copula densities, see \cite{huk2025your}).

\paragraph{Copula training}
We estimate the CDC parameters $\theta$ by minimizing a hybrid objective that combines dependence identification with score matching. The loss function is a weighted sum of a cross-entropy term ($\mathcal{L}_{CE}$) and a denoising score matching term ($\mathcal{L}_{MSE}$):
\begin{equation}
    \mathcal{L}_{c_{dc}}(\theta) = \alpha \sum_{s=1}^{K} \mathbb{E}_{z \sim \tilde{p}_{T_s}} \left[ -\log c_{dc}^{(s)}(z; \theta) \right] + \sum_{s=1}^{K} \mathbb{E}_{z_{T_1}, \epsilon} \left[ || \hat{\epsilon}_s(z_s; \theta) - \epsilon ||^2 \right]
\end{equation}
where $c_{dc}^{(s)}$ is the predicted probability of the input belonging to diffusion time step $s$, and $\hat{\epsilon}_s$ is the score-based noise prediction derived from the classifier's gradients. The first term ($\mathcal{L}_{CE}$) forces the model to learn the correct density ratio for likelihood estimation and calibration, while the second term ($\mathcal{L}_{MSE}$) ensures the gradients of the density are accurate, facilitating stable score-based sampling of joint tail events. Empirical advantages of this loss function are discussed in \cite{yadin2024classification, huk2025diffusion}.

\subsection{Sampling}
Future trajectories are generated via a two-stage procedure: sampling the dependence structure from the CDC, followed by mapping to the asset values using inverse marginal CDF transformations.

\textbf{Copula sampling.} We generate dependent samples using Langevin dynamics on the time augmented space \citep{song2019generative}. Starting from Gaussian noise $z_{T_K} \sim \mathcal{N}(0, I_d)$, we iteratively denoise samples using the approximate score function derived from the classifier gradients at times $T_k\leq s\leq 0$:
\begin{equation}
    \nabla_u \log c_s(u) \propto \nabla_z \log \mathbb{P}(t=T_s \mid z) - \nabla_z \log \mathbb{P}(t=T_K \mid z)
\end{equation}
Upon reaching $s=0$, the final Gaussian vector $z_0$ is transformed to the $[0,1]^d$ copula scale by marginally applying the inverse Gaussian distribution function $u^i = \Phi(z_0^i)$. This produces joint samples on the hypercube with the dependence structure of the $d$ assets. 

\textbf{Inverse Marginal Transformation.} To obtain predicted returns $y_t$, we apply the inverse Probability Integral Transform (PIT) marginally to copula samples $u$. Each uniform dimension $u^i$ is mapped through the inverse CDF of the Mixture Density Network, conditioned on the history $\mathcal{X}_t$:
\begin{equation}
    y_t^i = F^{-1}_i(u^i \mid \mathcal{X}_t; \Psi)
\end{equation}
where $F_i$ represents the cumulative distribution function of the weighted Normal, Laplace, and Student-t mixture defined in the predictive marginal model in Section~\ref{sec:pred_marg} above. As $u^i$ is uniform, the produced sample $y^i_t$ will exactly follow the predictive distribution $F_i$, thereby inheriting marginal statistical guarantees such as heavy tails.

\section{Relation to Prior Work}
Generative modeling for financial time series has increasingly embraced Diffusion Probabilistic Models, which offer superior stability and distributional fidelity compared to earlier adversarial approaches~\citep{MR4982120,shabsigh2023generative}. State-of-the-art methods such as  those of ~\cite{li2024transformer} and~\cite{tashiro2021csdi} treat multivariate forecasting as a unified generative task, learning the joint distribution of all assets simultaneously via a shared score function.

However, these end-to-end approaches face a critical trade-off. By optimizing a single joint objective, they often compromise the calibration of individual assets, failing to capture the precise volatility dynamics of specific instruments. Furthermore, relying on implicit dependence learning makes it difficult to rigorously parametrise complex tail associations. Crucially, our approach differs from standard diffusion-based time series models by explicitly isolating the dependence structure.

We further differ from works utilising copulas for time series \citep{patton2009copula,patton2012review,ashoktactis,czado2022vine,huk2023probabilistic} since our copula is modelled via a diffusion process instead of rigid parametric families. This lets us leverage the flexibility of powerful modern architectures for complex and multi-modal distributions, overcoming the brittleness of existing copula-based approaches to financial time series.

\section{Experiments}
\label{sec:exp}
In this section, we aim to address the following questions regarding financial forecasting relevance:

\textbf{Q1.} \textit{Are separate marginal models advantageous for calibration compared to joint diffusions?}

\textbf{Q2.} \textit{Is our method better suited for forecasting joint extreme events, such as crashes and bubbles?}

\paragraph{Evaluation Benchmarks.} As our approach is based on diffusion models, we benchmark against two leading methods for multivariate probabilistic time series forecasting that share this foundation: the Conditional score-based diffusion model of \cite{tashiro2021csdi} (CSDI) and the Transformer-modulated diffusion model of \cite{li2024transformer} (TMDM).

\paragraph{Metrics.} To answer \textbf{Q1}, we present Probability Integral Transformation (PIT, see Appendix~\ref{PIT})  in Figure \ref{fig:pit} and Quantile-Quantile (QQ, see Appendix~\ref{QQ}) plots in Figure \ref{fig:qq} as calibration diagnostics. We evaluate the Root Mean Squared Error (RMSE, see Appendix \ref{RMSE}), Mean Absolute Error (MAE, see Appendix \ref{MAE}), and Tail Event Accuracy (Tail, see Appendix \ref{Tail}) as pointwise metrics of forecast accuracy in Table \ref{tab:marginal}. Lastly, we compute the Continuous Ranked Probability Score (CRPS, see Appendix \ref{CRPS}) in Table \ref{tab:crps} to quantify the discrepancy between the generated forecast distributions and observed values. To answer \textbf{Q2}, we assess the models' capability to anticipate extreme market volatility. We analyse the correlation strength at extreme quantiles in Figure \ref{fig:corr} and the probability assigned to simultaneous joint extremes in Figure \ref{fig:jointX}. Finally, we evaluate robustness against ``black swan'' events in Figure \ref{fig:black_swan_comparison} and calculate the CRPS for joint extreme values in Table \ref{tab:crps}.

\begin{figure}[t]
    \centering
    \includegraphics[width=0.8\linewidth]{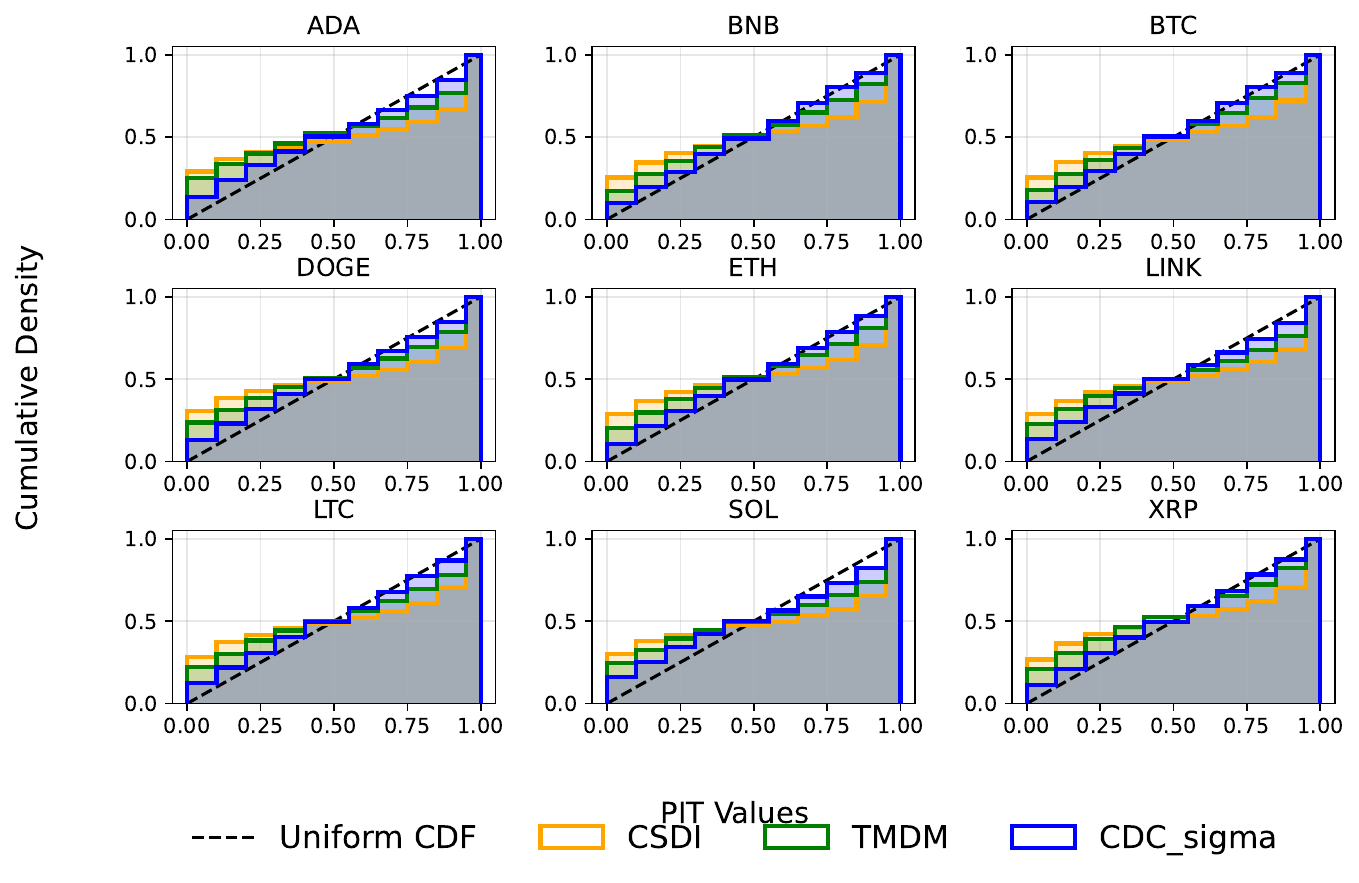}
    \caption{\textbf{Marginal Calibration.} Cumulative distribution of PIT values across nine assets. The black dashed line ($y=x$) represents perfect calibration, to which our model (blue)  closely adheres.}
    \label{fig:pit}
\end{figure}

\subsection{Marginal Calibration}
Effective marginal calibration ensures that the predicted distribution of each asset matches the empirical frequency and magnitude of returns, particularly in the tails. This alignment is critical, as tail events exert a dominant influence on risk management outcomes. 
\paragraph{General calibration.} We present the marginal PIT cumulative distributions in Figure \ref{fig:pit}. Our model closely adheres to the ideal diagonal line. Conversely, CSDI and TMDM display deviations near 0 and 1, indicative of predictive distributions with tails that are too thin compared to observed data. 

\begin{figure}
    \centering
    \includegraphics[width=0.7\linewidth]{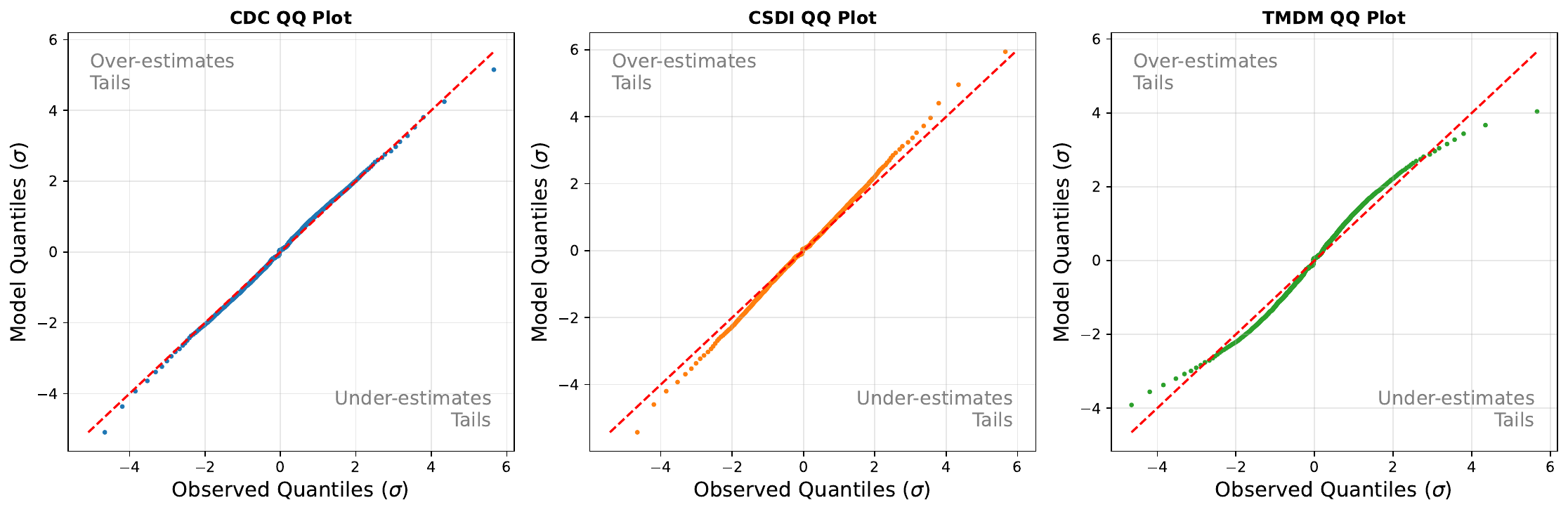}
    \caption{\textbf{Quantile-Quantile Distributional Diagnostics.} Points deviating from the red diagonal (perfect forecast) indicate a mismatch in distribution. The "S shape" of CSDI and TMDM indicates an underestimation of heavy tails, while CDC maintains alignment even in extreme quantiles.}
    \label{fig:qq}
\end{figure}

\paragraph{Tail behaviour.} To further scrutinise tail behaviour, we examine the QQ plots in Figure \ref{fig:qq}. Our model exhibits a robust match to empirical quantiles, following the theoretical diagonal, while CSDI and TMDM show noticeable discrepancies in the extreme regions. Finally, Table \ref{tab:marginal} reports RMSE, MAE, and CRPS as metrics of central accuracy, alongside Tail Event Accuracy. Our model secures the best results for RMSE and Tail Event Accuracy and ranks a close second on MAE and CRPS, indicating a favourable balance between capturing central tendency and calibrating for tail risk. This demonstrates the advantage of separate marginal models.

\begin{table}
  \caption{Comparison of marginal predictive performance across all models. Bold values indicate the best performance for each metric (lowest for RMSE, MAE, CRPS; highest for Tail).}
  \label{tab:marginal}
  \centering
  \begin{tabular}{lcccc}
    \toprule
    \textbf{Model} & \textbf{RMSE} $\downarrow$& \textbf{MAE} $\downarrow$& \textbf{CRPS} $\downarrow$& \textbf{Tail$>$0.005} $\uparrow$\\
    \midrule
    \textbf{CDC}  & \textbf{0.003137} & 0.002155 & 0.001756 & \textbf{0.025172} \\
    \textbf{CSDI} & 0.003140 & \textbf{0.002141} & \textbf{0.001643} & 0.015926 \\
    \textbf{TMDM} & 0.003225 & 0.002234 & 0.001662 & 0.004483 \\
    \bottomrule
  \end{tabular}
\end{table}

\subsection{Forecasting Joint Extremes}
Effective financial modeling must capture not only individual asset volatility but also the contagion effects where correlations converge during distress.

\begin{figure}[h]
    \centering
    \includegraphics[width=0.55\linewidth]{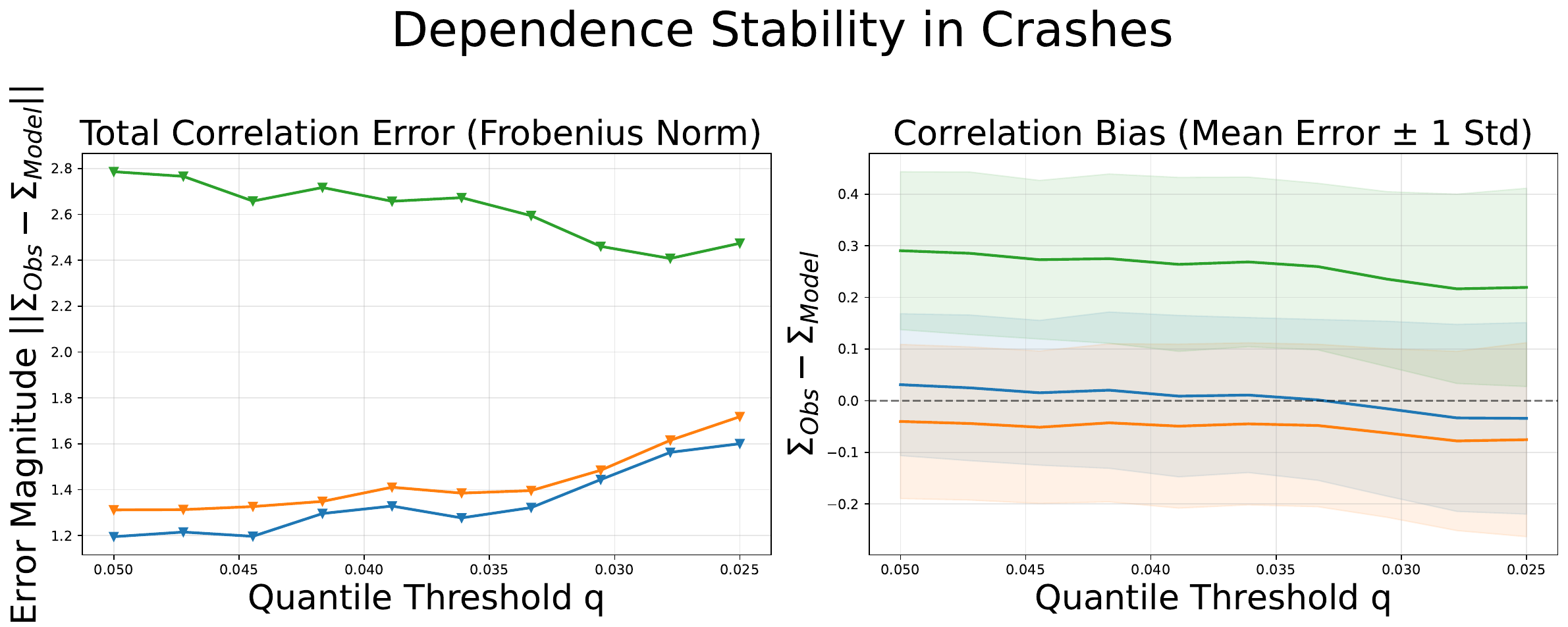}
         \includegraphics[width=0.55\linewidth]{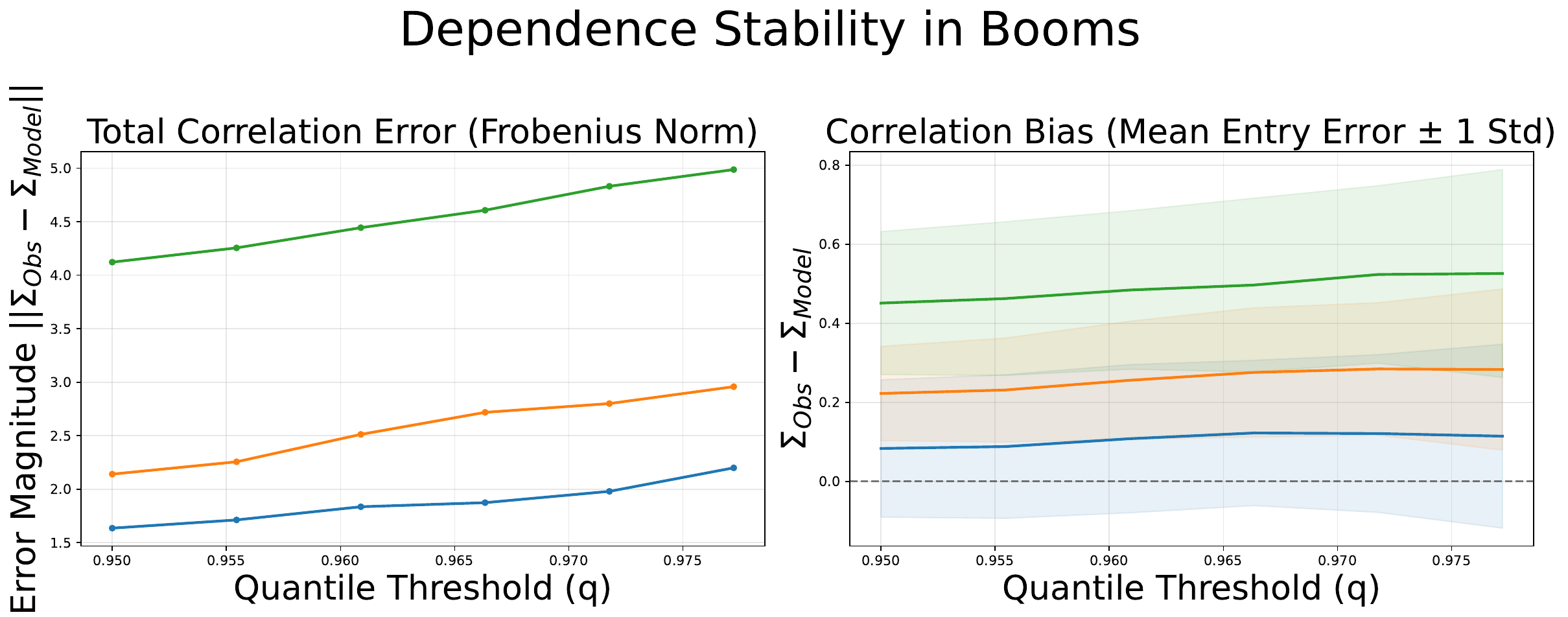}
   \includegraphics[width=0.55\linewidth]{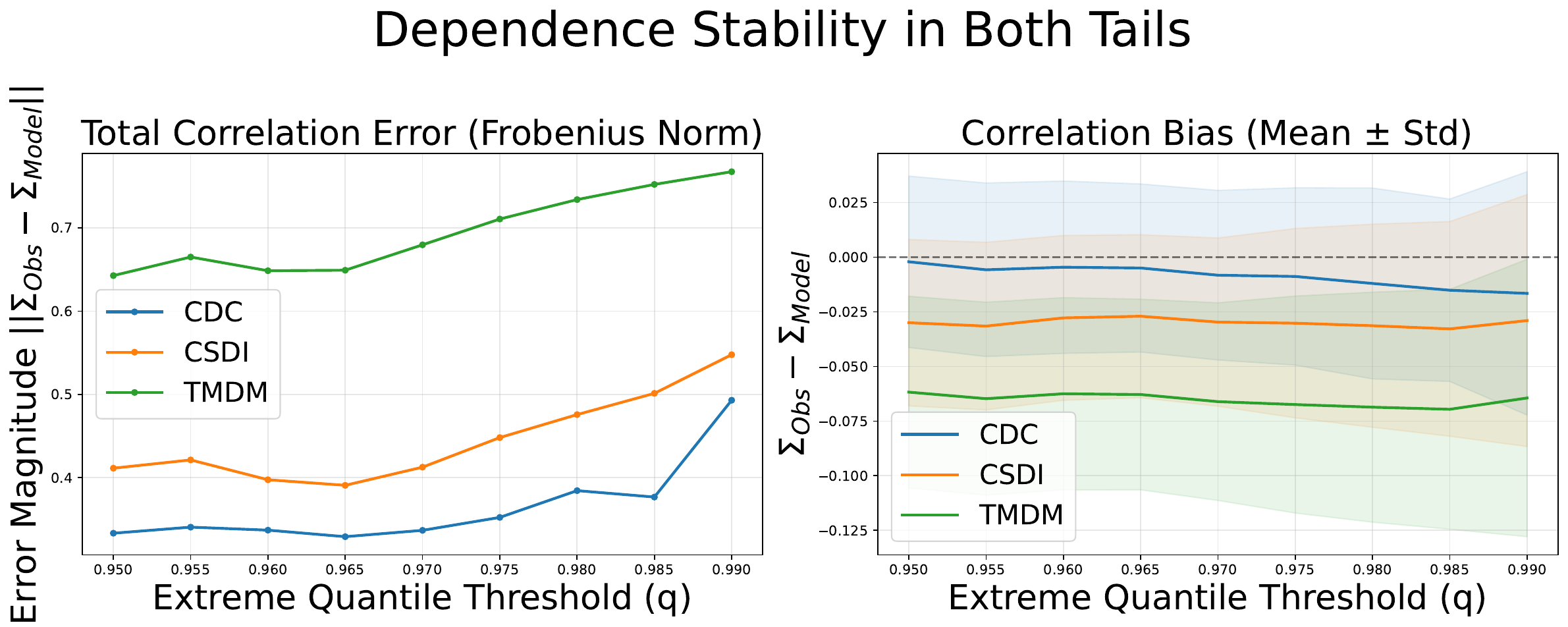}
 \caption{\textbf{Correlation structure of extremes.} Evaluation of correlation matrix accuracy as the quantile threshold \(q\) becomes more extreme. The left column of each figure reports the total error (Frobenius norm \(\| \Sigma_{Obs} - \Sigma_{Model} \|\)), while the right column tracks the mean bias. The CDC (blue) demonstrates superior stability and converging bias in the deep tail compared to CSDI (orange) and TMDM (green).}
    \label{fig:corr}
\end{figure}

\paragraph{Dependence Stability.} Figure \ref{fig:corr} evaluates correlation matrix accuracy as market conditions worsen. While CSDI (orange) performs well in moderate regimes ($q=0.10$), its bias drifts negative in deep tails ($q \to 0.03$), indicating an overestimation of correlation during crashes. In contrast, \textbf{CDC} (blue) demonstrates superior stability, with bias converging toward zero as events become more extreme, whereas TMDM (green) consistently fails to capture the heightened co-movement of assets.

\begin{figure}
    \centering
    \includegraphics[width=0.4\linewidth]{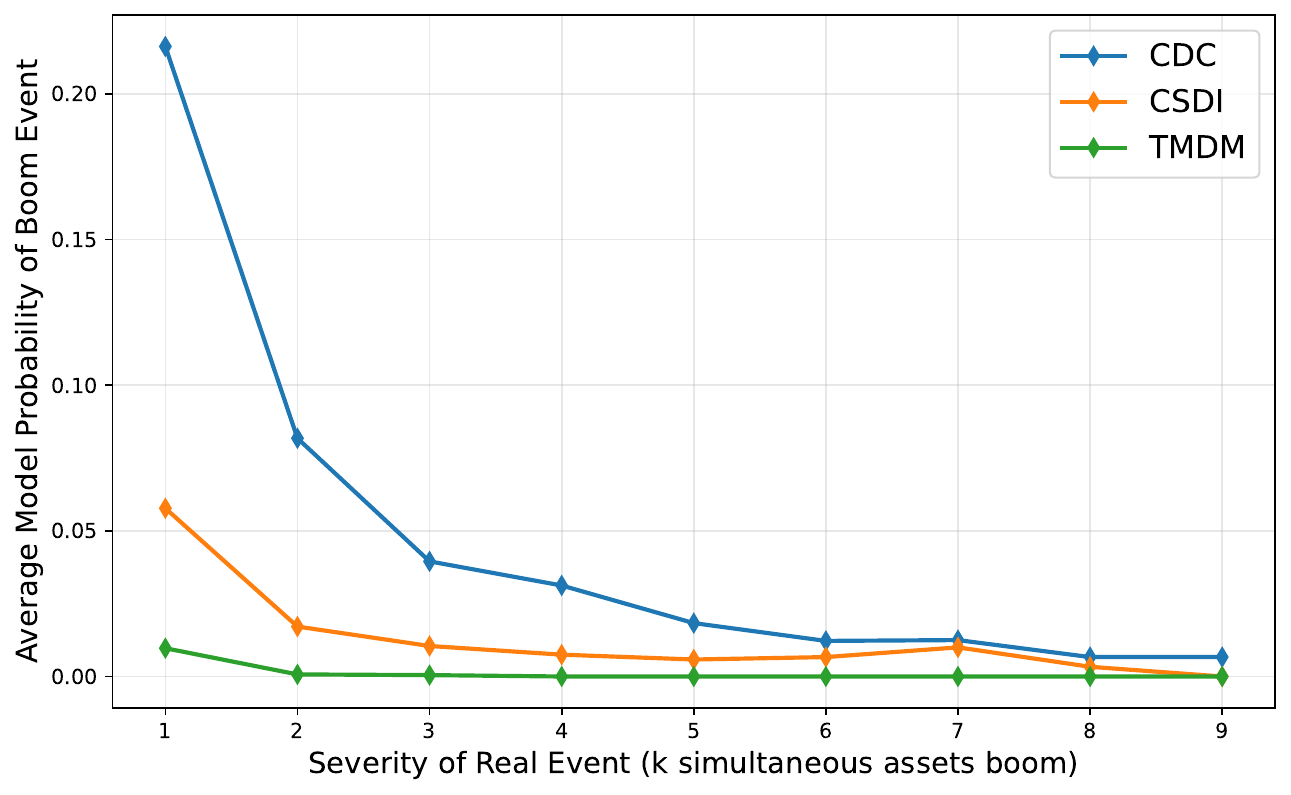}
     \includegraphics[width=0.4\linewidth]{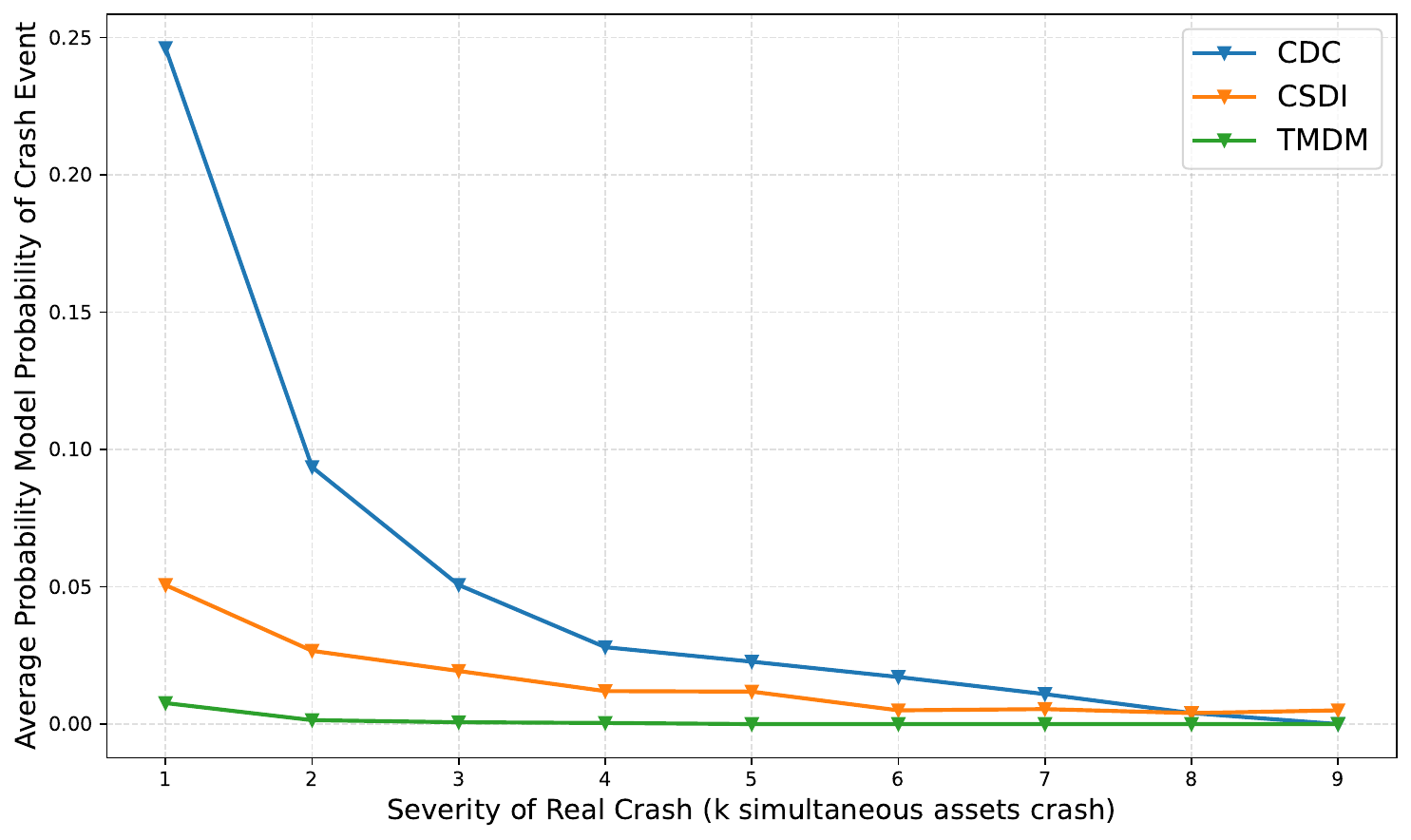}
    \caption{\textbf{Systemic Event Probability.} Average model-assigned probability of observing a simultaneous extreme event involving exactly $k$ assets. The x-axis represents the severity of the actual market event (number of assets $k$ simultaneously crashing/booming).}
    \label{fig:jointX}
\end{figure}

\paragraph{Systemic Event Probability.} In Figure \ref{fig:jointX}, we measure the probability assigned to simultaneous extreme returns across $k$ assets. While baseline models (CSDI, TMDM) exhibit rapid probability decay, effectively treating systemic crashes ($k \ge 4$) as statistical impossibilities, \textbf{CDC} maintains a significant probability buffer. This confirms its ability to anticipate the synchronized downturns characteristic of crypto markets.

\begin{wrapfigure}{r}{0.3\textwidth}
    \centering
    \vspace{-10pt}
    \caption{CRPS values for multiple assets in $5\%$ tails.}
    \begin{tabular}{l c}
        \hline
        \textbf{Model} & \textbf{CRPS} $\downarrow$ \\
        \hline
        \textbf{CDC}  & \textbf{0.00322}  \\
        \textbf{CSDI} & 0.00361 \\
        \textbf{TMDM} & 0.00372 \\
        \hline
    \end{tabular}
    \label{tab:crps}
\end{wrapfigure}

\begin{figure}
    \centering
    \includegraphics[width=0.65\linewidth]{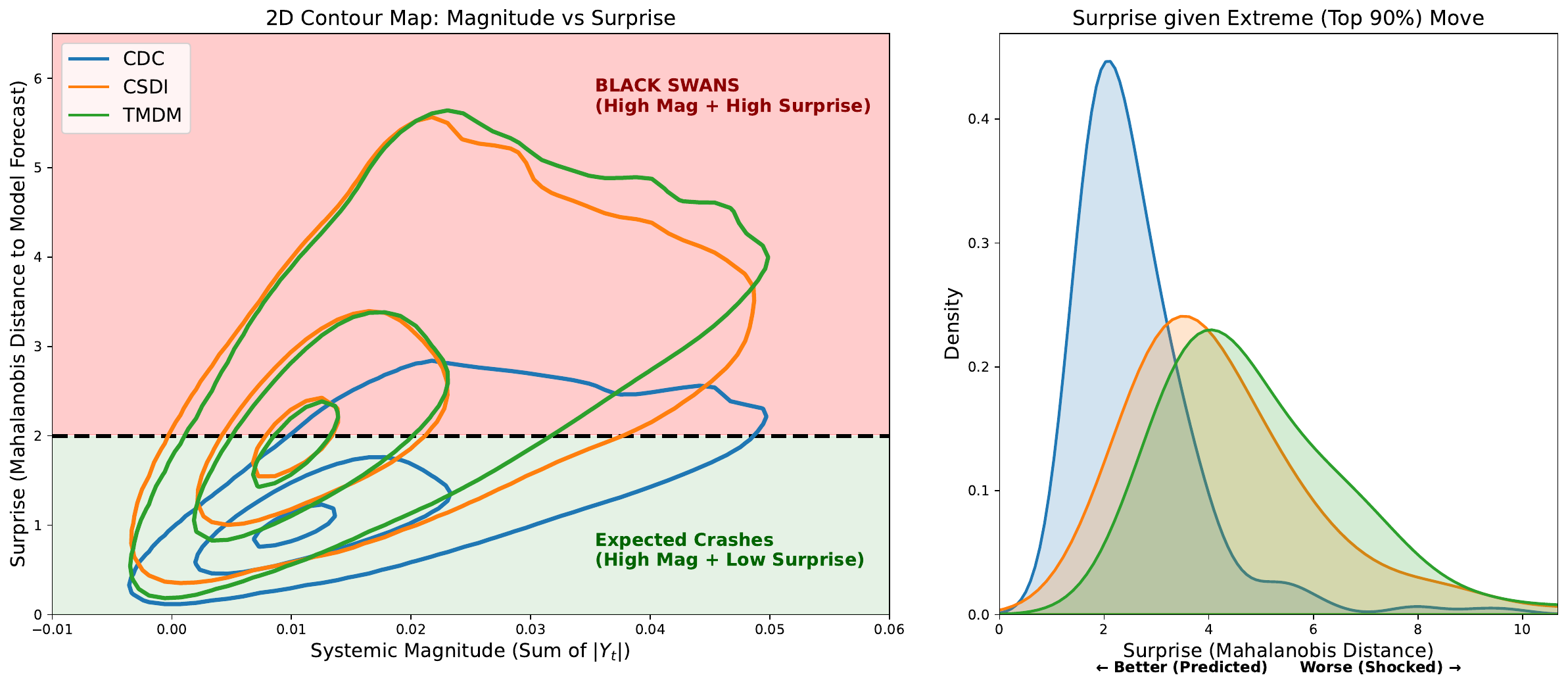}
    \caption{\textbf{From Black Swans to Expected Crashes.} A 2D contour map visualizing the relationship between Event Magnitude (sum of absolute returns, X-axis) and Model Surprise (Mahalanobis distance, Y-axis). The shaded regions divide the space into "Expected Crashes" (green, low surprise) and "Black Swans" (red, high surprise). %The contours reveal a critical divergence: as systemic magnitude increases ($X \to 0.05$), CSDI (orange) and TMDM (green) contours extend vertically into the high-surprise zone, indicating these models view such events as statistical anomalies. In contrast, CDC (blue) contours expand horizontally within the low-surprise zone, demonstrating that it correctly anticipates extreme volatility rather than being "shocked" by it.
    }
    \label{fig:black_swan_comparison}
\end{figure}

\paragraph{Black Swan Detection.} Figure \ref{fig:black_swan_comparison} visualizes the relationship between event magnitude and model surprise (Mahalanobis distance). The contours reveal a critical divergence: baselines classify high-magnitude events as "Black Swans" (high surprise, red zone), whereas \textbf{CDC} contours stretch horizontally, classifying them as "Expected Crashes" (low surprise, green zone). This geometric evidence proves that CDC effectively "de-risks" extreme volatility by incorporating it into its predictive distribution.

\paragraph{Quantitative Validation.} We also compute the CRPS on timesetps with more than two assets in either side of their $5\%$ tail as a
quantitative metric in  Table \ref{tab:crps}. Our model achieves the lowest CRPS on days with simultaneous extreme events, confirming its robustness in modeling joint tail risks.

\section{Discussion}
We proposed a Diffusion-Copula framework to reconcile the generative power of diffusion models with the strict calibration required for financial risk assessment. Our results highlight a fundamental trade-off: while end-to-end baselines (CSDI, TMDM) excel at capturing average-case dynamics, they exhibit a dangerous ``normality bias'' that erases extreme events. By decoupling the marginals, our model achieves superior heavy-tail calibration (Table \ref{tab:marginal}, Figures \ref{fig:pit}, \ref{fig:qq}) without sacrificing dependence accuracy. Crucially, the CDC is the only model capable of anticipating systemic crashes; whereas baselines classify simultaneous downturns as statistically impossible ``Black Swans'' (Figure \ref{fig:black_swan_comparison}), our approach correctly models the asymmetric correlation spikes characteristic of market contagion (Figure \ref{fig:jointX}).

\subsection{Future Directions}
While our novel framework demonstrates  promise, particularly in improving tail risk modelling, it remains a two-stage estimator where marginal errors inherently propagate to the dependence structure. A critical improvement could be achieved by employing a flexible, model-free approach for the marginals, such as Neural Spline Flows~\citep{durkan2019neural}, which would enhance robustness against parametric misspecification. Furthermore, the standard CDC training objective effectively learns densities but may not be the optimal architecture for all high-dimensional dependence structures. Future research could address these constraints by employing {Flow-based models}~\citep{lipman2022flow} to capture tail dependence with greater precision.

Practically, accurate modelling of \textit{joint tail risks} has profound financial implications. Unlike static Gaussian models, our framework captures asymmetric contagion—where correlations spike during crashes—enabling the construction of "tail-robust" portfolios. The generative nature of the diffusion process further empowers risk managers to perform {synthetic stress testing}, simulating plausible "black swan" scenarios essential for robust Expected Shortfall (ES) estimation and the pricing of complex multi-asset derivatives.

%%%%%%%%%%%%%%%%%%%%%%%%%%%%%%%%%%%%%%%%%%%%%%%%% End of main paper

%\subsubsection*{Author Contributions}
%If you'd like to, you may include  a section for author contributions as is done
%in many journals. This is optional and at the discretion of the authors.

%\subsubsection*{Acknowledgments}
%Use unnumbered third level headings for the acknowledgments. All
%acknowledgments, including those to funding agencies, go at the end of the paper.

\bibliography{main}
\bibliographystyle{iclr2026_conference}

\appendix

\section{Datasets}
Our data set in~\cite{CryptoDataDownload} comprises high-frequency OHLCV(Open, High, Low, Close, and Volume) data for nine prominent cryptocurrencies, including BTC, ETH, LTC, XRP, BNB, ADA, SOL, DOGE and LINK, recorded at one-minute resolution throughout 2022. To account for the inherent seasonality, cyclicality, and frequent volatility spikes characteristic of the crypto market, we downsampled the raw data to a 10-minute frequency. Our analysis specifically focuses on the first three quarters of 2022 to ensure consistency in market regimes.

\section{Experimental Details}
\label{apdx:exp}

In our experiments, we first independently model marginal distributions using a Mixture Density Network (MDN) comprising nine components (three Normal, three Laplace, and three Student’s t-distributions). To capture temporal dependencies under a 14-order Markov assumption, we utilize a dual-branch architecture: an LSTM processes historical return sequences with 128 hidden size and 5 layers, while a parallel MLP encodes auxiliary features, including volatility and path-dependent geometric metrics (e.g., path length and trend strength). To promote diverse sampling patterns and prevent mode collapse, we apply temperature scaling ($\tau = 0.5$) and entropy regularization ($\lambda = 0.01$) to the mixture weights.The framework is implemented in PyTorch and optimized using Adam with a weight decay of $10^{-5}$ and a OneCycle learning rate scheduler (peak $LR = 3 \times 10^{-5}$). To enhance throughput, we leverage automatic mixed-precision (AMP) training. 

Secondly, we model the dependence using the Classification-Diffusion Copula (CDC)~\citep{huk2025diffusion}. Within this framework, we evaluate three specific models: a CDC model with an identity matrix in the Ornstein-Uhlenbeck (OU) process, a CDC model with a full covariance matrix in the OU process, and a flow matching model.

All experiments (including the benchmark below) were conducted on an NVIDIA RTX 4090D GPU.

\section{Benchmark}

\textbf{CSDI}. We adopt the imputation component from the CSDI~\citep{tashiro2021csdi}, which is achieved via a masking strategy. The prediction problem can thus be viewed as a missing data imputation problem, where the model infers the 15th value based on the 14 preceding observed data points. In our model, the default batch size is set to 512. We employ a reverse process in conditional diffusion model with $T=50$ sampling steps, and at each timestep, we draw 100 Monte Carlo samples.

\textbf{TMDM}. Transformer-Modulated Diffusion Models~\citep{li2024transformer} introduces an enhanced Transformer architecture optimized to handle the challenges of non-stationarity in temporal data. The task is formulated as a sequence-to-sequence forecasting problem where the model predicts the next return $r_{t+1}$ given a historical window of length $L=14$ ($r_{t-13}, \dots, r_t$). The architecture employs an Encoder-Decoder framework: the Encoder processes the full 14-step historical sequence to extract global temporal dependencies, while the Decoder uses the last 7 steps as a 'start token' to guide the autoregressive generation of the next step. To balance sampling diversity with computational efficiency, we set the number of iterations for the reverse process to $T=100$, with 100 samples generated at each timestep. We employ a batch size of 64 for training and 512 for testing, while the patience for early stopping is set to 10.

\section{Diagnostics and Metrics}

\subsection{Metrics}
\subsubsection{Root Mean Square Error (RMSE) }\label{RMSE} RMSE is a standard metric used to evaluate the accuracy of a predictive model. It measures the average magnitude of error between predicted values and observed values. The formula is:

\[
\text{RMSE} = \sqrt{\frac{1}{n} \sum_{i=1}^{n} (y_i - \hat{y}_i)^2}
\]

where \(y_i\) is the observed value, \(\hat{y}_i\) is the predicted value, and \(n\) is the number of observations. 

\subsubsection{Mean Absolute Error (MAE) }\label{MAE} MAE measures the average absolute difference between predicted and observed values. The formula is:

\[
\text{MAE} = \frac{1}{n} \sum_{i=1}^{n} |y_i - \hat{y}_i|
\]

where \(y_i\) is the observed value, \(\hat{y}_i\) is the predicted value, and \(n\) is the number of observations. 

\subsubsection{Continuous Ranked Probability Score (CRPS)}\label{CRPS}CRPS measures the difference between the predicted cumulative distribution function (CDF) and the observed value (a step function). The formula is:

\[
\text{CRPS}(F, y) = \int_{-\infty}^{\infty} \left[ F(x) - \mathbf{1}\{x \geq y\} \right]^2 dx
\]

where \(F\) is the predicted CDF, \(y\) is the observed value, and \(\mathbf{1}\) is the indicator function. CRPS generalizes the Mean Absolute Error (MAE) to probabilistic forecasts.

\subsubsection{Tail Event Accuracy (Tail)}\label{Tail}
The Tail metric focuses on the model's ability to predict extreme market movements. It isolates "tail events," defined as instances where the observed return exceeds a specific threshold \( \tau \) (e.g., \( x\% \)). The metric measures the proportion of these actual tail events that the model correctly predicted would also exceed the threshold. The formula is:

\[
\text{Tail} = \frac{\sum_{i=1}^{n} \mathbf{1}\{y_i > \tau \land \hat{y}_i > \tau\}}{\sum_{i=1}^{n} \mathbf{1}\{y_i > \tau\}}
\]

where \(y_i\) is the observed value, \(\hat{y}_i\) is the predicted value, \(\tau\) is the defined percentage threshold, and \(\mathbf{1}\) is the indicator function (taking the value 1 if the condition is true, and 0 otherwise). This effectively calculates the recall rate for extreme positive jumps.

\subsubsection{Probability Integral Transformation (PIT)}\label{PIT}
The Probability Integral Transformation (PIT) is a diagnostic tool used to assess the calibration of predictive distributions. It relies on the principle that if the predicted cumulative distribution function (CDF), \(\hat{F}_t\), matches the true data-generating process, the evaluation of the observed values under this distribution should be uniformly distributed. The PIT value \(u_t\) for an observation \(y_t\) is calculated as:

\[
u_t = \hat{F}_t(y_t) = \int_{-\infty}^{y_t} \hat{p}_t(x) dx
\]

where \(\hat{p}_t\) is the predicted probability density function. If the model is well-calibrated, the sequence \(u_t\) follows a Uniform distribution \(U[0, 1]\). Visually, a histogram of \(u_t\) values should be flat; deviations (e.g., U-shaped or hump-shaped histograms) indicate under-dispersion or over-dispersion in the model's uncertainty estimates.

\subsubsection{Quantile-Quantile Plot (QQ)}\label{QQ}
The Quantile-Quantile (QQ) plot is a graphical method for comparing the distributional shape of the model's standardized residuals (or transformed PIT values) against a theoretical reference distribution (typically the Standard Normal distribution). It plots the quantiles of the empirical data against the quantiles of the theoretical distribution. The coordinates \((x_i, y_i)\) for the plot are given by:

\[
(x_i, y_i) = \left( \Phi^{-1}\left(\frac{i}{n+1}\right), z_{(i)} \right)
\]

where \(z_{(i)}\) represents the \(i\)-th ordered standardized residual (or probit-transformed PIT value), \(n\) is the sample size, and \(\Phi^{-1}\) is the inverse CDF of the standard normal distribution. Points falling on the \(45^\circ\) line indicate that the model's distributional assumptions (e.g., tail behavior and skewness) align with the observed data.
% \textbf{CRPS-sum}. CRPS summed over multiple locations (CRPS-sum) is an extension of the CRPS used to evaluate multivariate or spatio-temporal probabilistic forecasts. It aggregates the CRPS across all target dimensions. The common formula for a forecast over \(L\) locations is:

% \[
% \text{CRPS\_sum} = \sum_{l=1}^{L} \text{CRPS}(F_l, y_l)
% \]

% where \(F_l\) is the predictive CDF and \(y_l\) is the observation at location \(l\). This provides a single score summarizing the overall performance across the field. It is particularly useful for verifying ensemble predictions of spatial fields or multi-step time series, balancing the assessment of both marginal calibration and dependency structure across dimensions.

\subsection{Diagnostics}
\textbf{Black swan plot}. The black swan plot is a diagnostic visualization tool used in probabilistic forecasting and risk assessment to evaluate a model's performance in predicting extreme, low-probability events, often referred to as "black swan" events, which focuses on the tails of the predictive distribution.

For a set of probabilistic forecasts, the observed outcomes are ranked against the corresponding predictive distributions. The plot then displays the frequency or proportion of observations that fall within increasingly extreme quantile intervals of the forecasts (e.g., beyond the 95th, 99th, or 99.9th percentiles). 

Interpretation:

If the curve aligns with the diagonal line of perfect calibration, the model accurately captures the likelihood of extreme events.

If the curve lies above the diagonal, the model underestimates tail risk (observations are more extreme than predicted).

If the curve lies below the diagonal, the model overestimates tail risk (observations are less extreme than predicted).

\end{document}